\documentclass{article}
\usepackage{ijcai23}

\usepackage{times}
\usepackage{soul}
\usepackage{url}
\usepackage[hidelinks]{hyperref}
\usepackage[utf8]{inputenc}
\usepackage[small]{caption}
\usepackage{graphicx}
\usepackage{amsmath}
\usepackage{amsthm}
\usepackage{booktabs}
\usepackage{algorithm}
\usepackage{algorithmic}
\usepackage[switch]{lineno}

\usepackage{amssymb}
\usepackage{subfig}
\usepackage{rotating}
\usepackage{multirow}
\usepackage{multicol}
\usepackage{amsfonts}       
\usepackage{nicefrac}       
\usepackage{microtype}      
\usepackage{xcolor}         

\newtheorem{definition}{Definition}[]

\newcommand{\Con}{\operatorname{\textbf{Con}}}

\title{ Geometric Relational Embeddings: A Survey\footnote{Work in progress} }

\author{
Bo Xiong$^1$\and
Mojtaba Nayyeri$^1$\and
Ming Jin$^{2}$\and
Yunjie He$^{1}$\and\\
Michael Cochez$^{3}$\and
Shirui Pan$^{4}$\and
Steffen Staab$^{1,5}$
\affiliations
$^1$University of Stuttgart, $^2$Monash University, $^3$VU Amsterdam\\
$^4$Griffith University,  $^5$University of Southampton\\
\emails
}

\begin{document}
\maketitle

\begin{abstract}
Geometric relational embeddings map relational data as geometric objects that combine vector information suitable for machine learning and structured/relational information for structured/relational reasoning, typically in low dimensions. Their preservation of relational structures and their appealing properties and interpretability have led to their uptake for tasks such as knowledge graph completion, ontology and hierarchy reasoning, logical query answering, and hierarchical multi-label classification.
We survey  methods that underly geometric relational embeddings and categorize them based on (i) the embedding geometries that are used to represent the data; and (ii) the relational reasoning tasks that they aim to improve. We identify the desired properties (i.e., inductive biases) of each kind of embedding and discuss some potential future work. 
\end{abstract}

\section{Introduction}

Vector embeddings map objects to a low-dimensional vector space. Vector embeddings
have been developed to learn representations of objects that allow for distinguishing relevant differences and ignoring irrelevant variations between objects. 
When vector embeddings are used to embed structural/relational data, they fail to represent key properties of relational data
 that cannot be easily modeled in a plain, low-dimensional vector space.  For example, relational data may have been defined by applying set operators such as set inclusion and exclusion \cite{xiong2022faithful}, logical operations such as negation \cite{ren2020beta}, or they may exhibit relational patterns like the symmetry of relations \cite{abboud2020boxe} and structural patterns (e.g., trees and cycles) \cite{chami2020low,xiong2022ultrahyperbolic}.

\begin{figure}
    \centering
    \subfloat[\centering Ball embedding]{{\includegraphics[width=0.28\linewidth]{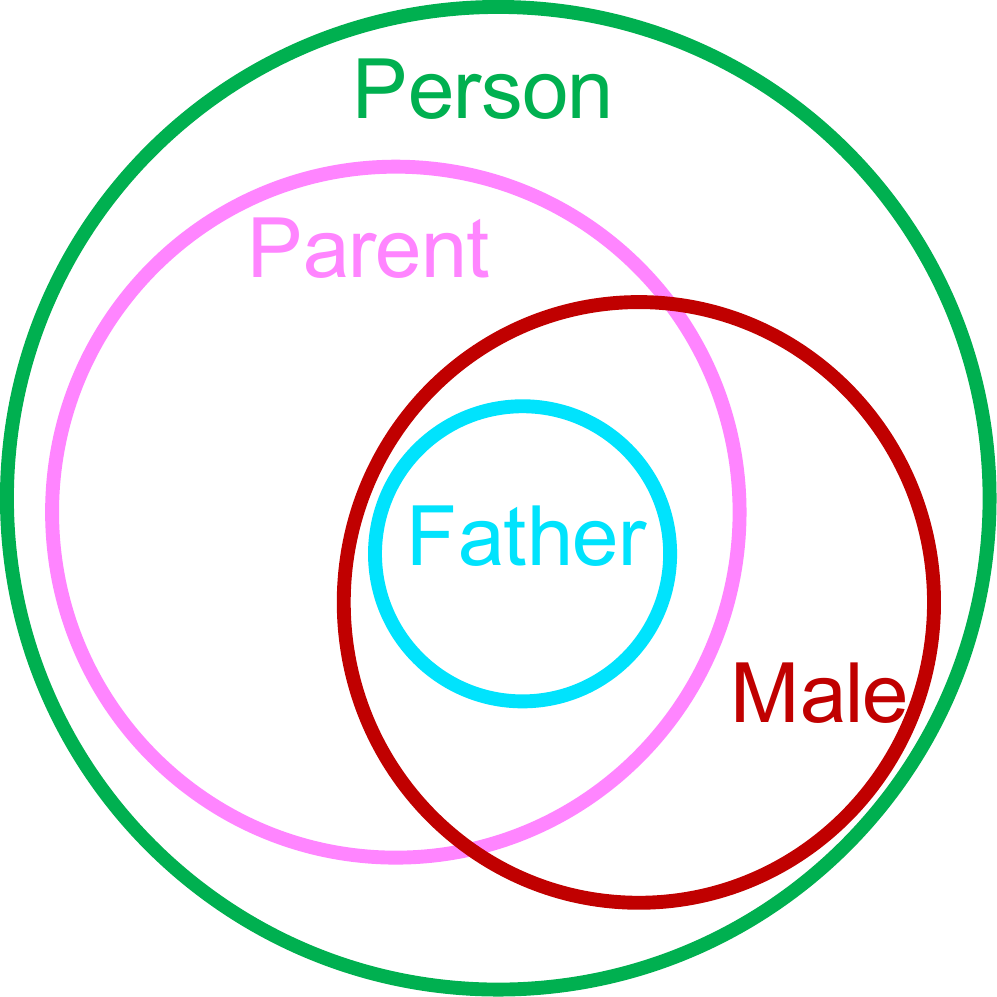} }}%
    \quad 
    \subfloat[\centering Box embedding]{{\includegraphics[width=0.29\linewidth]{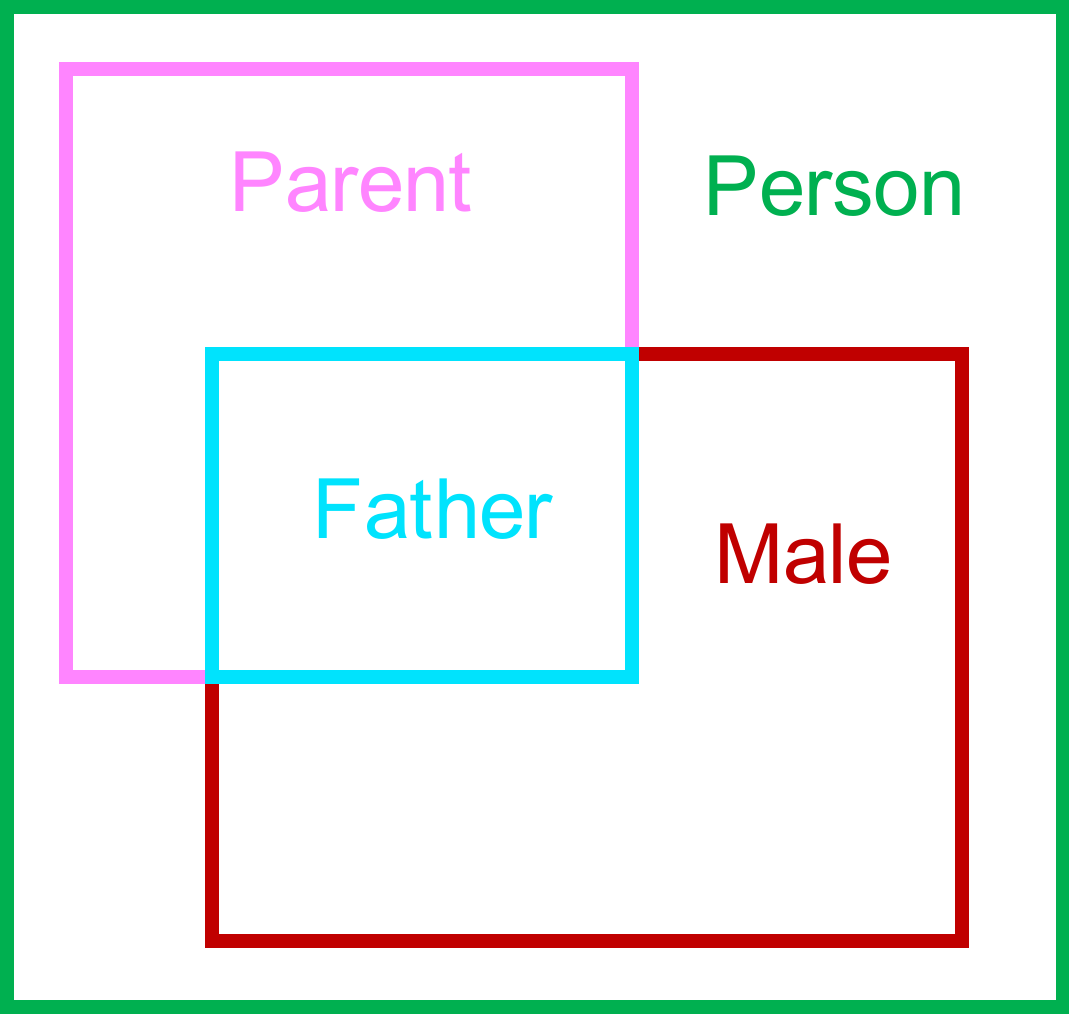} }}%
    \quad 
    \subfloat[\centering Cone embedding]
    {{\includegraphics[width=0.29\linewidth]{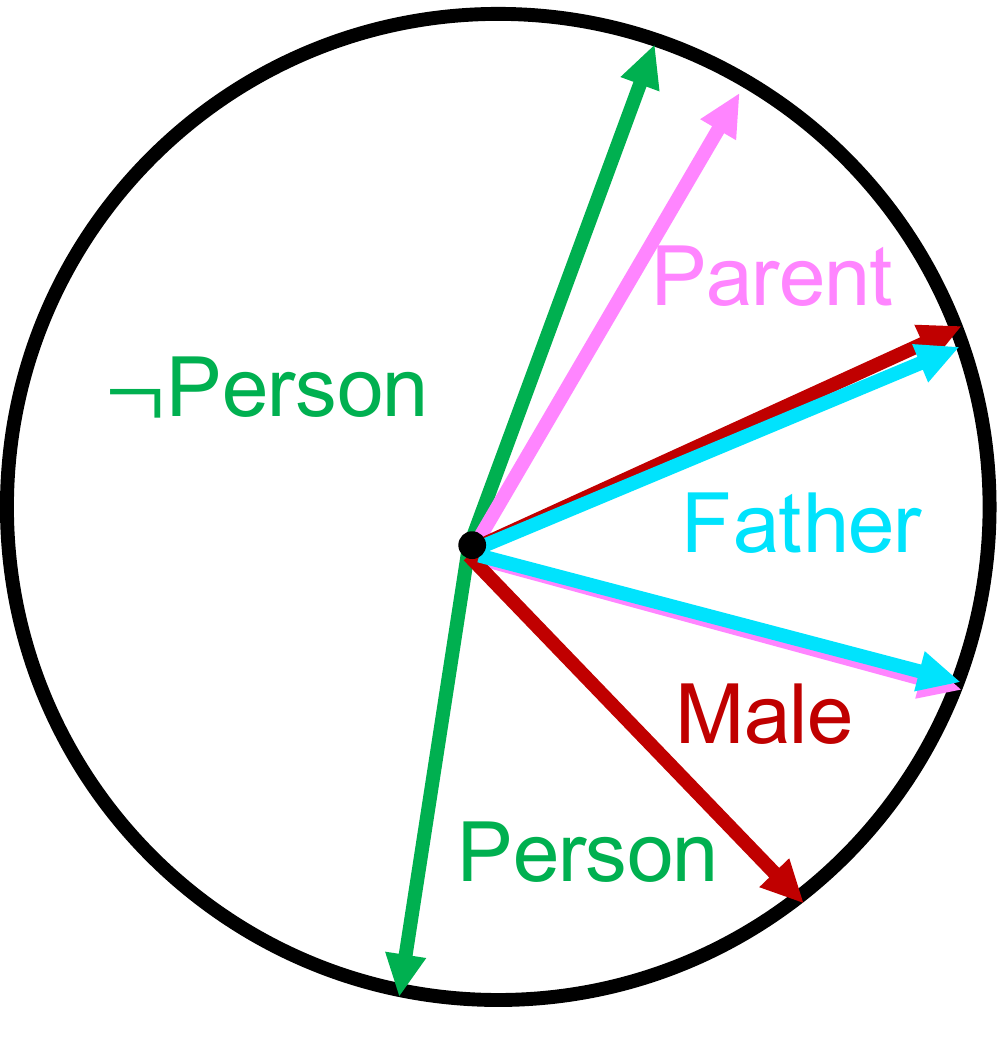} }}%
    \caption{An illustration of region-based embeddings. (a) Ball embedding represents concepts as $n$-ball allowing for modeling set inclusion; (b) Box embedding represents concepts as boxes that further support intersectional closure; (c) Cone embedding represents concepts as cones that support intersectional and negation closures.}
    \label{fig:region-based}%
    \vspace{-0.5cm}
\end{figure}

Going beyond plain vector embeddings, \emph{geometric relational embeddings} replace the vector representations with more advanced geometric objects, such as convex regions \cite{DBLP:conf/ijcai/KulmanovLYH19,ren2019query2box,xiong2022faithful}, density functions \cite{wang2022dirie,ren2020beta}, elements of hyperbolic manifolds \cite{chami2020low,xiong2022ultrahyperbolic}, and their combinations \cite{suzuki2019hyperbolic}. 
Geometric relational embeddings provide a rich geometric inductive bias for modeling relational/structured data. 
For example, embedding objects as convex regions allows for modeling not only similarity but also set-based and logical operators such as set inclusion, set intersection \cite{xiong2022faithful} and logical negation \cite{zhang2021cone} (cf. Fig. \ref{fig:region-based}) while representing data on non-Euclidean manifolds allows for capturing complex structural patterns, such as representing hierarchies in hyperbolic space \cite{chami2020low}. 

Geometric relational embeddings have been successfully applied in many relational reasoning tasks including but not limited to knowledge graph (KG) completion \cite{abboud2020boxe,xiong2022ultrahyperbolic},  ontology/hierarchy reasoning \cite{vilnis2018probabilistic,xiong2022faithful}, hierarchical multi-label classification (HMC) \cite{patel2021modeling,xionghyperbolic}, and logical query answering \cite{ren2020beta}.
Different downstream applications require varying capabilities from underlying embeddings and, hence, appropriate choice requires a sufficiently precise understanding of embeddings' characteristics and task requirements.

\begin{figure*}
    \centering
    \includegraphics[width=0.85\textwidth]{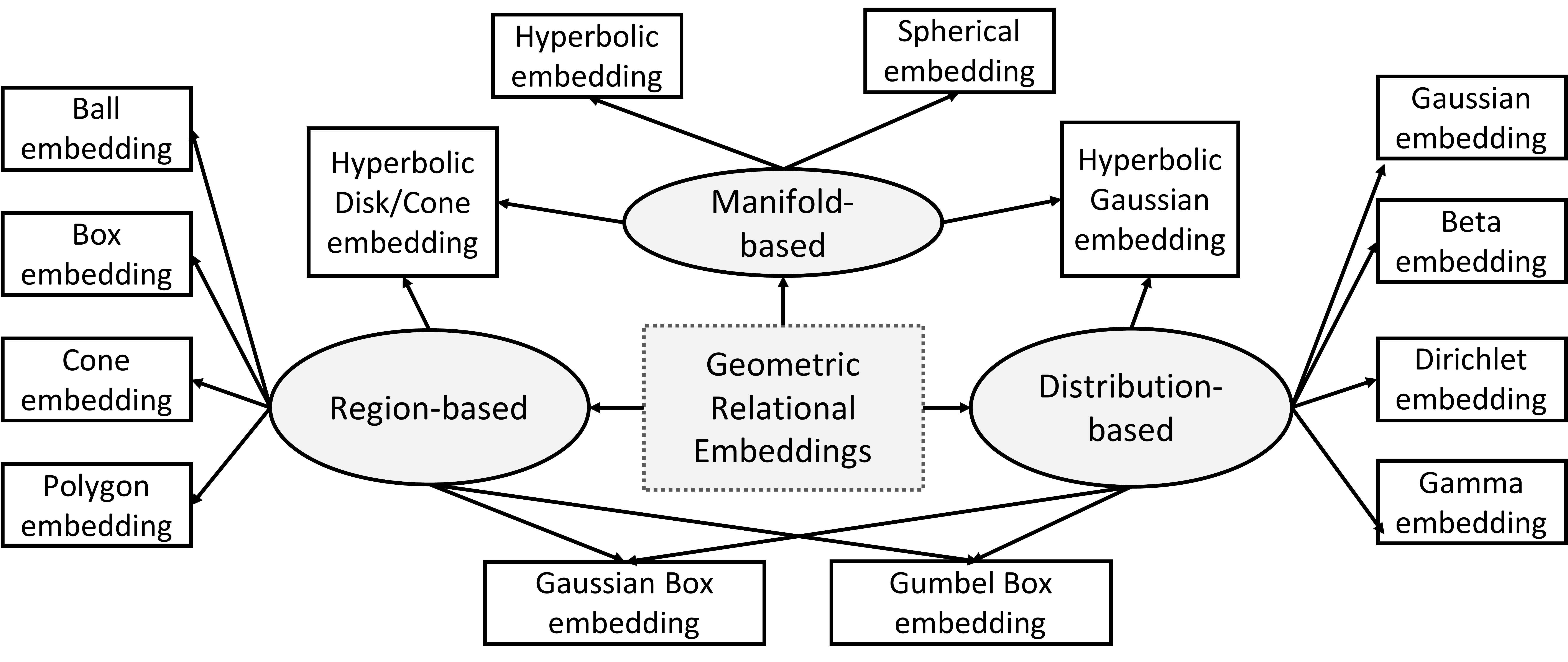}
    \caption{A systematic categorization of geometric relational embeddings. The methods for geometric relational embeddings can be classified into distribution-based, region-based, and manifold-based. Each two of them can be combined and form a hybrid method, e.g., the hyperbolic cone is a combination of hyperbolic (manifold-based) and cone embedding (region-based). 
}
    \label{fig:taxinomy_method}

\end{figure*}

In this paper, we  identify the desired capabilities of geometric relational embeddings on different relational reasoning tasks. 
We  conduct a comprehensive survey on the methods of geometric relational embeddings. 
The methods are classified based on the underlying geometries used to represent objects as well as the targeted relational reasoning tasks. 
Fig. \ref{fig:taxinomy_method} shows a taxonomy of these methods. 
We systematically discuss and compare the inductive biases of different embeddings and discuss the applications in four relational reasoning tasks: KG completion, ontology and hierarchy completion, hierarchical multi-label classification, and logical query answering. 

\section{Preliminaries and Problem Definition}

We describe input relational data as a directed edge-labeled graph or multi-relational graph \cite{hogan2021knowledge}. 

\begin{definition}[directed edge-labeled graph]
A directed edge-labeled graph is a tuple $G=(V, E, L)$, where $V \subseteq \Con$ is a set of nodes, $L \subseteq \Con$ is a set of edge labels, and $E \subseteq V \times L \times V$ is a set of edges, where $\Con$ denote a countably infinite set of constants.
\end{definition}

Such data structures can be used to represent a knowledge graph. For example, in a KG about proteins, nodes represent proteins and edges represent (binary) interactions between proteins. Using reserved keywords, such as defined for RDFS or OWL \cite{hitzler2009foundations}, directed edge-labeled graphs can also be used to define ontologies. E.g., in the gene ontology, nodes represent gene functional concepts and edges represent (binary) relations between them. Ontology-based KGs comprise facts about entities and complex concepts definitions. 


Classical logical reasoning on structured knowledge remains unchallenged when it comes to sound and complete deduction of new facts from given, consistent relational data. 
Geometric relational embeddings operate on the premise that relational data is incomplete and some missing facts must be derived by induction or transduction. 
To this end, all geometric relational embeddings establish principles of analogy. Embeddings of relations are sought such that all the pairs of nodes in a relation are located in analogous manner to each other in geometric space. Then, key functional capabilities of geometric relational embeddings include the querying for similar nodes and the completion of triple patterns in the absence of directly matching triples. If a triple $(h,r,t)\notin E$ is plausible, then queries in form of the triple patterns $(h,r,?t)$ or $(h,?r,t)$ should bind the query variable $?t$ to $t$  or the query variable $?r$ to $r$, respectively.

\begin{table}[]
    \centering
    \resizebox{1.03\linewidth}{!}{
    \begin{tabular}{ccccccc}
\hline
& properties & hierarchy & KG & ontology & HMC & query   \\
\hline
similarity & & $\checkmark$ & $\checkmark$ & $\checkmark$ & $\checkmark$ & $\checkmark$ \\
\hline
uncertainty & & $\checkmark$ & $\checkmark$ & $\checkmark$ & $\checkmark$ & $\checkmark$ \\
\hline
\multirow{3}{*}{\begin{turn}{-0}set theory\end{turn} } 
& inclusion & $\times$ & $\times$ & $\checkmark$ & $\checkmark$ & $\checkmark$   \\
& exclusion & $\times$ & $\times$ & $\checkmark$ & $\checkmark$ & $\checkmark$ \\
& overlap & $\times$ & $\times$ & $\checkmark$ & $\checkmark$ & $\checkmark$ \\
& difference & $\times$ & $\times$ & $\checkmark$ & $\checkmark$ & $\checkmark$ \\
\hline
\multirow{3}{*}{\begin{turn}{-0}logical\end{turn}} 
& intersection & $\times$ & $\times$ & $\checkmark$ & $\times$ & $\checkmark$ \\
& union & $\times$ & $\times$ & $\checkmark$ & $\times$ & $\checkmark$\\
& negation & $\times$ & $\times$ & $\checkmark$ & $\times$ & $\checkmark$ \\
\hline
\multirow{3}{*}{\begin{turn}{-0}relational\end{turn}} 
& symmetry & $\times$ &  $\checkmark$ & $\checkmark$ & $\times$ & $\checkmark$\\
& inversion & $\times$ &  $\checkmark$ & $\checkmark$ & $\times$ & $\checkmark$\\
& composition & $\times$ & $\checkmark$ & $\checkmark$ & $\times$ & $\checkmark$\\
& transitivity  & $\checkmark$ & $\checkmark$ & $\checkmark$ & $\checkmark$ & $\checkmark$ \\
\hline
\multirow{3}{*}{\begin{turn}{-0}structural\end{turn}} 
& cycles & $\times$ &  $\checkmark$ & $\times$ & $\times$ & $\checkmark$ \\
& trees & $\checkmark$ &  $\checkmark$ & $\checkmark$ & $\checkmark$ & $\checkmark$\\
& tree-cycle & $\checkmark$ &  $\checkmark$ & $\checkmark$ & $\times$ & $\checkmark$ \\
\hline
\end{tabular}
}
\caption{The desiderata of different relational reasoning tasks. }
\label{tab:comparision}
\end{table}

\begin{table*}[ht!]
    \centering
    \resizebox{\textwidth}{!}{
    \begin{tabular}{cccccccccc}
\hline
Category & Method & Embedding & Task & Institute & Venue & Year \\
\hline
\multirow{9}{*}{\begin{turn}{-0}Distribution-based\end{turn}} 
& KG2E \cite{he2015learning} & Gauss & KG & CAS & CIKM & 2015\\
& TransG \cite{xiao2016transg} & Gauss & KG & Tsinghua & ACL & 2016\\
& HCGE \cite{santos2016multilabel} & Gauss & HMC & Sorbonne & ECML PKDD & 2016\\
& PERM \cite{choudhary2021probabilistic} & Gauss & query &  Virginia Tec & NeurIPS & 2021\\
& BetaE \cite{ren2020beta} & Beta & query & Stanford & NeurIPS & 2020\\
& DiriE \cite{wang2022dirie} & Dirichlet & KG & BUPT & WWW & 2022\\
& GammE \cite{yang2022gammae} & Gamma & query & OPPO & EMNLP & 2022\\
\hline
\multirow{17}{*}{\begin{turn}{-0}Region-based\end{turn}} 
& ELEm \cite{DBLP:conf/ijcai/KulmanovLYH19} & Ball & ontology & KAUST & IJCAI & 2019 \\
& \cite{DBLP:conf/iclr/DongBJLCSCZ19} & Ball & hierarchy & U. Bonn & ICLR & 2019\\
& Box Lattice \cite{vilnis2018probabilistic} & Box & hierarchy & UMass & ACL & 2018 \\
& Joint Box \cite{patel2020representing} & Box &  hierarchy &  UMass &  AKBC & 2020\\
& BEUrRE \cite{DBLP:conf/naacl/ChenBCDLM21} & Box & KG &  UCLA &  NAACL-HLT & 2021\\
& BoxE \cite{abboud2020boxe} & Box & KG &  Oxford &  NeurIPS & 2021\\
& Box4Type \cite{DBLP:conf/acl/OnoeBMD20} & Box & HMC & UT Austin &  ACL & 2021\\
& Quer2Box \cite{ren2019query2box} & Box & query & Stanford &  ICLR & 2019\\
& MBM \cite{patel2021modeling} & Box & HMC & UMass &  ICLR & 2022\\
& BoxEL \cite{xiong2022faithful} & Box & ontology & USTUTT &  ISWC & 2022\\
& BC-BOX \cite{zhangmodeling} & Box & hierarchy & UMass &  NeurIPS & 2022\\
& OE \cite{vendrov2015order} & Cone & hierarchy & U. Toronto & ICLR & 2016\\
& Cone Semmantic \cite{o2021cone} & Cone & ontology & U. Lubeck & IJCAI & 2021\\
& ConE \cite{zhang2021cone} & Cone & query & CAS & NeurIPS & 2021\\
& ExpressE \cite{DBLP:journals/corr/abs-2206-04192} & Polygons & KGs & TU Vienna & ICLR & 2023\\
\hline
\multirow{16}{*}{\begin{turn}{-0}Manifold-based\end{turn}} 
& 5$*$E \cite{nayyeri20215} & Spherical  & KG & U. Bonn & AAAI & 2020 \\
& MuRS \cite{wang2021mixed} & Spherical & KG & UIUC & WWW & 2021 \\
& MuRP \cite{balazevic2019multi} & Hyperbolic & KG & Edinburgh & NeurIPS & 2019 \\
& AttH \cite{chami2020low} & Hyperbolic & KG & Stanford & ACL & 2020 \\
& HyperIM \cite{DBLP:conf/aaai/ChenHXCJ20} & Hyperbolic & HMC & BJTU & AAAI & 2020 \\
& HBE \cite{pan2021hyperbolic} & Hyperbolic & KG & Southeast & EMNLP & 2021 \\
& HyboNet \cite{chen2021fully} & Hyperbolic & KG & Tsinghua & ACL & 2022 \\
& HMI \cite{xionghyperbolic} & Hyperbolic & HMC & USTUTT & NeurIPS & 2022 \\
& MuRMP \cite{wang2021mixed} & Mixed & KG & UIUC & WWW & 2021 \\
& GIE \cite{cao2022geometry} & Mixed & KG & CAS & EMNLP & 2022 \\
& UltraE \cite{xiong2022ultrahyperbolic} & Mixed & KG & USTUTT & KDD & 2022 \\
& DGS \cite{iyer2022dual} & Mixed & KG & UCLA & KDD & 2022 \\
\hline
\multirow{10}{*}{\begin{turn}{-0}Hybrid\end{turn}} 
& Smooth Box \cite{DBLP:conf/iclr/LiVZBM19} & Gauss+Box & hierarchy & UMass & ICLR & 2019 \\
& Gumbel Box \cite{dasgupta2020improving} & Gumbel+Box & hierarchy & UMass & NeurIPS & 2020 \\
& HWN \cite{nagano2019wrapped} & Hyperbolic+Gauss & hierarchy & U. Tokyo & ICML & 2019\\
& ROWN \cite{cho2022rotated} & Hyperbolic+Gauss & hierarchy & POSTECH & NeurIPS & 2022\\
& HypDisk \cite{suzuki2019hyperbolic} & Hyperbolic+Ball & hierarchy & LAPRA & ICML & 2019\\
& HypE \cite{DBLP:conf/www/ChoudharyRKSR21} & Hyperbolic+Box & query &  Virginia Tech & WWW & 2021\\
& HypCone \cite{ganea2018hyperbolic}  & Hyperbolic+Cone & hierarchy & ETH & ICML & 2018\\
& ConeE \cite{bai2021modeling} & Hyperbolic+Cone & KG & Tsinghua & NeurIPS & 2021\\
& \cite{dhall2020hierarchical}  & Hyperbolic+Cone & HMC  & ETH & CVPR & 2020\\
& PolygonE \cite{DBLP:conf/aaai/YanZSXLLLW22} & Hyperbolic+Polygon & KG  & CAS & AAAI & 2022\\
\hline
    \end{tabular}
    }
\caption{Summary of the geometric relational embedding methods. HMC means hierarchical multi-label classification.}
  \label{tab:grl_methods}
\end{table*}

\paragraph{Desiderata.} 
The key functional capability of a geometric relational embedding model is based on the proper judgement of plausibility, whose quality benefits if relational properties and patterns can be represented geometrically in the embedding space as follows: 1) \textbf{similarity:} an embedding model should produce similar embeddings for similar entities/relations; 2) \textbf{uncertainty:} an embedding model should capture the \emph{degree of truthfulness}; 3) \textbf{set theory:} an embedding model should capture set-theoretical semantics including set inclusion, set exclusion, set overlap, and set difference; 4) \textbf{logical operations:} an embedding model should capture logical connectives such as logical conjunction, negation, and union. 
5) \textbf{relational patterns:} an embedding model should capture some relational patterns like symmetry, inversion, composition, and transitivity; and 6) \textbf{structural patterns:} an embedding model should capture structural/topological patterns such as trees, cycles, and their combination. 
The desiderata differ w.r.t different relational reasoning tasks. For example, logical operations only matter when dealing with ontologies, and queries with conjunction and negation might exist. 
Table \ref{tab:comparision} summarizes the desiderata of each relational reasoning task. 

\section{Geometric Relational Embeddings}

As Table \ref{tab:grl_methods} summarizes, the methodologies can be divided into four categories based on the underlying geometries.

\subsection{Distribution-based Embeddings}

Probability distributions provide a rich geometry of the latent space. Their density can be interpreted as soft regions and it allows us to model uncertainty, asymmetry, set inclusion/exclusion, entailment, and so on. 

\paragraph{Gaussian embeddings.}

Word2Gauss \cite{vilnis2015word} maps words to multi-dimensional Gaussian distributions over a latent embedding space such that the linguistic properties of the words are captured by the relationships between the distributions.
A Gaussian $\mathcal{N}(\boldsymbol{\mu}, \boldsymbol{\Sigma})$ is parameterized by a mean vector $\boldsymbol{\mu}$ and a covariance matrix $\boldsymbol{\Sigma}$ (usually a diagonal matrix for the sake of computing efficiency). 
The model can be optimized by an energy function $-E\left(\mathcal{N}_i, \mathcal{N}_j\right)$ that is equivalent to the KL-divergence $D_{\operatorname{KL}}\left(\mathcal{N}_j \| \mathcal{N}_i\right)$ defined as
\begin{equation}
D_{\operatorname{KL}}\left(\mathcal{N}_j \| \mathcal{N}_i\right) =\int_{x \in \mathbb{R}^n} \mathcal{N}\left(x ; \mu_i, \Sigma_i\right) \log \frac{\mathcal{N}\left(x ; \mu_j, \Sigma_j\right)}{\mathcal{N}\left(x ; \mu_i, \Sigma_i\right)} d x.  
\end{equation}

KG2E \cite{he2015learning} extends this idea to KG embedding by mapping entities and relations as Gaussians. 
Given a fact $(h,r,t)$, the scoring function is defined as $f(h, r, t)=\frac{1}{2}\left(\mathcal{D}_{\operatorname{KL}}\left(\mathcal{N}_h, \mathcal{N}_r\right)+\mathcal{D}_{\operatorname{KL}}\left(\mathcal{N}_r, \mathcal{N}_t\right)\right)$.
The covariances of entity and relation embeddings allow us to model uncertainties in KGs.
While modeling the scores of triples as KL-divergence allows us to capture asymmetry. 
TransG \cite{xiao2016transg} generalizes KG2E to a Gaussian mixture distribution to deal with multiple relation semantics revealed by the entity pairs.
For example, the relation HasPart has at least two latent semantics: composition-related as  (\emph{Table}, \emph{HasPart}, \emph{Leg}) and location-related as (\emph{Atlantics}, \emph{HasPart}, \emph{NewYorkBay}).


\paragraph{Dirichlet embeddings.}
DiriE \cite{wang2022dirie} embeds entities as Dirichlet distributions $f(\boldsymbol{x} | \boldsymbol{\alpha})=\frac{1}{\mathrm{~B}(\boldsymbol{\alpha})} \prod_{i=1}^d x_i^{\alpha_i-1}$ where $\boldsymbol{\alpha}=\left(\alpha_1, \alpha_2, \cdots, \alpha_d\right)>0$ is the distribution parameter, $\mathrm{B}(\boldsymbol{\alpha})=\frac{\prod_{i=1}^d \Gamma\left(\alpha_i\right)}{\Gamma\left(\sum_{i=1}^d \alpha_i\right)}$ and $\Gamma(\alpha)=\int_0^{\infty} t^{\alpha-1} e^{-t} d t$ are the multivariate beta function and Gamma function, respectively. DiriE embeds relations as multinomial distributions $f(\boldsymbol{x} | \boldsymbol{\mu})=\frac{\Gamma\left(\sum_i \mu_i+1\right)}{\prod_i \Gamma\left(\mu_i+1\right)} \prod_{i=1}^k x_i^{\mu_i}$ where $\mu_i \in \mathbb{R}^{+}$is the distribution parameter. Given a triple $(h,r,t)$, since the Dirichlet distribution is a conjugate prior of the multinomial distribution and according to Bayesian inference, DiriE views the head entity as a prior distribution, the tail entity as a posterior distribution and the relation as a likelihood function that transforms the prior to the posterior distribution, i.e., $q_t \propto b_r p_h $ and $q_h \propto b_{r^{-1}} p_t$,
where $p$ and $q$ are PDFs of Dirichlet distributions, and $b$ is PMF of the multinomial distribution. 
Due to the flexibility of the multinomial distributions of relation modeling, DiriE is able to model various relational patterns, including symmetry, inversion and composition, e.g., a symmetric relation can be modeled by learning $b_r$ as $b_{r^{-1}}$.

\paragraph{Beta embeddings.}
Logical query answering includes logical connectives such as negation which can be properly modeled by distributions.
BetaE \cite{ren2020beta} embeds entities and queries as multivariate beta distributions \(f(\mathbf{x}|\boldsymbol{\alpha},\boldsymbol{\beta}) = \frac{\mathbf{x}^{\boldsymbol{\alpha}-1}(1-\mathbf{x}^{{\boldsymbol{\beta}-1}})}{\mathbf{B}(\boldsymbol{\alpha},\boldsymbol{\beta})}\) and models the negation by taking the reciprocal of the parameters $\alpha$ and $\beta$. 
By doing so, BetaE converts the regions with high density into low-density areas and vice versa, which resembles the desired property of negation. In addition, a conjunction of entity sets $S_{1},\dots, S_{n}$ can be modeled as a weighted product of involved Beta distributions,i.e., $f_{\operatorname{inter}} = \frac{1}{Z}\prod f_{S_i}^{w_{i}}$,
where $Z$ is a normalization constant and $w_{1},\dots, w_{n}$ are the attention weights summing to 1. Such weighted product of Beta distribution follows the intuition of conjunction, namely the regions of high density in the intersection set should also have high density in all input distributions. 
 
\paragraph{Gamma embeddings.}
GammaE \cite{yang2022gammae} improves BetaE by embedding entities and queries as Gamma distributions defined as $f(x|\alpha, \beta)=\frac{x^{\alpha-1} e^{-\beta x} \beta^\alpha}{\Gamma(\alpha)}$, where $x>0, \alpha>0$ is the shape, $\beta>0$ is the rate, and $\Gamma(*)$ is the Gamma function. GammaE models intersection as product of Gamma distributions $P_{S_{\text {Inter }}}=\frac{1}{Z} \prod_{i=1}^k P_{S_i}^{w_i}$ and union as Gamma mixture models $P_{S_{\text {Union}}}=\sum_{i=1}^k \theta_i P_{S_i}$ where $\theta_i=\frac{\exp \left(\left(P_{S_i}\right)\right.}{\sum_j \exp \left(P_{S_i}\right)}$, and $P_{S_i}=f\left(x|\alpha_i, \beta_i\right)$. 
Similar to BetaE, GammaE models negation by reversing the shape of the density function $N_{P_S}=f\left(x|\frac{1}{\alpha}, \beta\right)+\epsilon$ where $P_S=f(x|\alpha, \beta)$ and $\epsilon \in(0,1)$ is the elasticity.

\subsection{Region-based Embeddings}

Mapping data as convex regions is inspired by the Venn diagram. Region-based embeddings nicely model the set theory that can be used to capture uncertainty \cite{DBLP:conf/naacl/ChenBCDLM21}, logical rules \cite{abboud2020boxe}, transitive closure~\cite{vilnis2018probabilistic}, logical operations~\cite{ren2019query2box}, etc. Several convex regions have been explored including balls, boxes, and cones. Fig. \ref{fig:region-based} shows examples of the three embeddings. 

\paragraph{Ball embeddings}

Ball embedding associates each object $w$ with an $n$-dimensional ball $\mathbb{B}_w\left(\mathbf{c}_w, r_w\right)$, where $\mathbf{c}_w$ and $r_w$ are the central point and its radius of the ball, respectively.
A ball is defined as the set of vectors whose Euclidean distance to $\mathbf{c}_w$ is less than $r_w$: $\mathbb{B}\left(\mathbf{c}_w, r_w\right) = \left\{\mathbf{p}| \left\|\mathbf{c}_w-\mathbf{p}\right\|<r_w\right\}$. 
In ElEm \cite{DBLP:conf/ijcai/KulmanovLYH19}, each concept in $\mathcal{E}\mathcal{L}^{++}$ ontologies is represented as an open $n$-ball, and subsumption relations between concepts are modeled as ball containment. This explicit modeling of subsumption structure leads to significant improvements in predicting human protein-protein interactions. \cite{DBLP:conf/iclr/DongBJLCSCZ19} represents categories with $n$-balls while considering tree-structured category information. By embedding subordinate relations between categories as ball containment, it shows promising results on NLP tasks compared to conventional word embeddings.

\paragraph{Box embeddings}
Box embeddings represent objects with $d$-dimensional rectangles, i.e., a Cartesian product of $d$ closed intervals denoted by $\prod_{i=1}^d\left[x_i^{\mathrm{m}}, x_i^{\mathrm{M}}\right], \quad$ where $x_i^{\mathrm{m}}<x_i^{\mathrm{M}}$ and $x_i^{\mathrm{m}}$ and $x_i^{\mathrm{M}}$ are lower-left and top-right coordinates of boxes \cite{vilnis2018probabilistic}.
Box embeddings capture anticorrelation and disjoint concepts better than order embeddings. 
Joint Box \cite{patel2020representing} improves box embeddings' ability to express multiple hierarchical relations (e.g., hypernymy and meronymy) and proposes joint embedding of these relations in the same subspace to enhance entity characterization, resulting in improved performance.
BoxE \cite{abboud2020boxe} proposes embedding entities and relations as points and boxes in knowledge bases, improving expressivity by modeling rich logic hierarchies in higher-arity relations. 
BEUrRE \cite{DBLP:conf/naacl/ChenBCDLM21} is similar to BoxE but it differs in embedding entities and relations as boxes and affine transformations, respectively, which enables better modeling marginal and joint entity probabilities. 
Query2Box \cite{ren2019query2box} shares BoxE's idea, embedding queries as boxes, with answer entities inside, to support a range of querying operations, including disjunctions, in large-scale knowledge graphs. 
BoxEL \cite{xiong2022faithful}, ELBE \cite{DBLP:journals/corr/abs-2202-14018} and Box$^2$EL \cite{jackermeier2023box} propose to capture concept-level knowledge in the description logic $\mathcal{E}\mathcal{L}^{++}$ by modeling concepts and/or roles as axis-aligned boxes such that concept and role inclusions can be geometrically captured.   

\paragraph{Cone embeddings}
Cone embedding was first proposed in Order embedding (OE) \cite{vendrov2015order} to represent a partial ordered set (poset). However, this method is restricted to axis-parallel cones and it only captures positive correlations as any two axis-parallel cones intersect at infinite.
Recent cone embeddings formulate cones with additional angle parameters. 
As one of the main advantages, angular cone embedding has been used to model the negation operator since the angular cone is closed under negation. 
For example, negation can be modeled as the polarity of a cone as used in ALC ontology \cite{o2021cone} or the complement of a cone as used in ConE for logical queries \cite{zhang2021cone}. 

\paragraph{Polygon embeddings}
PolygonE \cite{DBLP:conf/aaai/YanZSXLLLW22} models n-ary relational data as gyro-polygons in hyperbolic space,  where entities in facts are represented as vertexes of gyro-polygons and relations as entity translocation operations. ExpressivE \cite{DBLP:journals/corr/abs-2206-04192} embeds pairs of entities as points and relations as hyper-parallelograms in the virtual triple space $\mathbb{R}^{2d}$. Such more flexible modeling allows for not only capturing a rich set of relational patterns jointly but additionally to display any supported relational pattern through the spatial relation of hyper-parallelograms.

\subsection{Manifold-based Embeddings}
The ability of embedding models to capture complex structural patterns intrinsically bounded by the volume of the embedding space. 
This consequently leads to distortion issues during the embedding of relational data with complex structural patterns.
A thread of work tries to mitigate this problem for a certain class of patterns e.g., hierarchy by embedding on a non-Euclidean manifold which is a smooth geometric object $\mathcal{M}$ (e.g., Poincare ball) which is locally Euclidean, but not globally. 
A Riemannian manifold is a pair $(\mathcal{M}, g)$ where $g$ is a metric tensor used to define the distance between any two points on the manifold.
Several methods have been proposed to embed relational data on non-Euclidean manifolds such as Poincare balls and Sphere \cite{weber2018curvature} to model hierarchical and loop structures, respectively. 


\paragraph{Hyperbolic embeddings}
The Poincaré ball
has been used for modeling hierarchical structures \cite{weber2018curvature}.
MuRP \cite{balazevic2019multi} models multi-relational data by transforming entity embeddings by Möbius matrix-vector multiplication and addition.
To capture both hierarchy and logical patterns e.g., symmetry and composition,
AttH \cite{chami2020low} models relation transformation by rotation/reflection and also embeds entities on the Poincaré ball.
FieldH and FieldP \cite{nayyeri2021knowledgefieldemi} embed entities on trajectories that lie on hyperbolic manifolds namely Hyperboloid and Poincaré ball to capture heterogeneous patterns formed by a single relation (e.g., a combination of loop and path) besides hierarchy and logical patterns. 
\cite{pan2021hyperbolic} embeds nodes on the extended Poincaré  Ball and  polar coordinate system to address numerical issues when dealing with the embedding of neighboring nodes on the boundary of the ball on previous models. 
HyboNet \cite{chen2021fully} proposes a fully hyperbolic method that uses Lorentz transformations to overcome the incapability of previous hyperbolic methods of fully exploiting the advantages of hyperbolic space. 
Poincaré Glove \cite{tifrea2018poincare} adapts the GloVe algorithm to learn unsupervised embedding in a Cartesian product of hyperbolic spaces which is theoretically connected to the Gaussian word embeddings and their Fisher geometry.
The application of hyperbolic geometry in entity alignment has been done in \cite{sun2020knowledge} where HypKA embeds two graphs on a Poincaré ball and then aligns the corresponding node pairs in two graphs. 
HMI \cite{xionghyperbolic} targets the problem of structured multi-label classification where the labels are organized under implication and mutual exclusion constraints, and are encoded as Poincaré hyperplanes that work as linear decision boundaries.
In addition to the mentioned works, the application of hyperbolic manifolds in temporal knowledge graph embedding has been done in \cite{montella2021hyperbolictemporal} which is an extension of AttH \cite{chami2020low} with time-dependent curvature.

\paragraph{Spherical embeddings}
A spherical manifold $\mathcal{M} = \{x\in \mathbb{R}^d |\,\,\, \|x\|=1\}$ has been used as embedding space in several works to model loops in graphs. 
MuRS \cite{wang2021mixed} models relations as linear transformations on the tangent space of spherical manifold, together with spherical distance in score formulation.
A more sophisticated approach is FiledP \cite{nayyeri2021knowledgefieldemi} in which entity spaces are trajectories on a sphere based on relations to model more complex patterns compared to MuRS.
5*E \cite{nayyeri20215} embeds entities on flows on the product space of a complex projective line which is called the Riemann sphere. 
The projective transformation then covers translating, rotation, homothety, reflection, and inversion which are powerful for modeling various structures and patterns such as combination of loop and loop, loop and path, and two connected path structures.

\paragraph{Mixed manifold embeddings}
Relational data exhibit heterogeneous structures and patterns where each class of patterns and structures can be modeled efficiently by a particular manifold.
Therefore, combining several manifolds for embedding is beneficial and addressed in the literature.
MuRMP \cite{wang2021mixed} extends MuRP \cite{balazevic2019multi} to a product space of spherical, Euclidean, and hyperbolic manifolds. 
GIE \cite{cao2022geometry} improves the previous mixed models by computing the geometric interaction on tangent space of Euclidean, spherical and hyperbolic manifolds via an attention mechanism to emphasize on the most relevant geometry and then projects back the obtained vector to the hyperbolic manifold for score calculation. 
DGS \cite{iyer2022dual} targets the same problem of heterogeneity of structures and utilizes the hyperbolic space, spherical space, and intersecting space in a unified framework for learning embeddings of different portions of two-view KGs (ontology and instance levels) in different geometric space.
While the reviewed works provide separate spherical, Euclidean, and hyperbolic manifolds and act on Riemannian manifolds, 
UltraE \cite{xiong2022ultrahyperbolic} embeds graphs on the Ultrahyperbolic manifold which is a semi-Riemannian manifold that seamlessly interleaves Euclidean, hyperbolic and spherical manifolds. 
In this model, relations are modeled by the pseudo orthogonal transformation which is decomposed into cosine-sine forms covering hyperbolic/circular rotation and reflection. This model can capture heterogeneous structures and logical patterns.

\subsection{Hybrid Embeddings}
Many methods try to combine the advantages of using distributions/regions and the advantages of using manifolds. Classical box embeddings use a hard volume in the objective function, leading to gradient issues. 
Soft volume-based methods like smoothed box \cite{DBLP:conf/iclr/LiVZBM19} and Gumbel box \cite{dasgupta2020improving} alleviate this problem by modeling the box via a Gaussian process and Gumbel process, respectively.
To simultaneously model hierarchies and set-theoretic semantics, Hyperbolic disk embedding \cite{suzuki2019hyperbolic} embeds relational objects as disks defined in hyperbolic space. 
HypE \cite{DBLP:conf/www/ChoudharyRKSR21}, on the other hand, simulates the box in hyperbolic space, but, this is no longer closed under intersection. A successful combination of manifolds and region-based methods is the hyperbolic cone embedding \cite{ganea2018hyperbolic} that models relational objects as angular cones in hyperbolic space.
ConE\cite{bai2021modeling} further extends it to logical queries on KGs. 
To combine the advantages of distribution-based methods and manifolds, HWN \cite{nagano2019wrapped} and RoHWN \cite{cho2022rotated} generalize Gaussian distributions to hyperbolic space, namely hyperbolic wrapped Gaussian.

\section{Applications in Relational Reasoning}

\subsection{Knowledge Graph Completion.}
To perform KG completion, 
a KGE model takes a triple pattern $(h,r,?)$ as a query, 
and replaces ``?" by an entities in the KG to generate a new triple $(h,r,e)$. 
Table \ref{tab:characteristics_KGE} summarizes the characteristics of some prominent geometric KG embeddings. 

To capture the uncertainty in a KG, KG2E \cite{he2015learning} embeds each entity and relation as a Gaussian distribution parameterized by a mean vector and diagonal covariance matrix. 
KG2E measures the plausibility of the triple by $\mathcal{D}_{\mathcal{K} \mathcal{L}}\left(\mathcal{P}_e, \mathcal{P}_r\right)$, where $\mathcal{P}_e \sim \mathcal{N}\left(\boldsymbol{\mu}_h-\boldsymbol{\mu}_t, \boldsymbol{\Sigma}_h+\boldsymbol{\Sigma}_t\right)$ and $\mathcal{D}_{\mathcal{K} \mathcal{L}}$ is the KL divergence. 
TransG \cite{xiao2016transg} generalizes the idea of KG2E to a mixture of Gaussian distribution to deal with the issue of multiple relation semantics.
DiriE models entities as Dirichlet distributions and relations as multinomial distributions, which is able to model relational patterns.


BEUrRE \cite{DBLP:conf/naacl/ChenBCDLM21} models uncertainty of KGs by using box embedding, where entities are mapped as boxes (axis-parallel rectangulars) and relations are modeled as affine transformations involving a translation of the position of boxes and a scaling of the size of boxes. BoxE, on the other hand, embeds entities as points, and relations as a set of boxes with the number of boxes depending on the arity of the relation. Hence, BoxE is able to model n-ary relations. Besides, the relational boxes spatially characterize the basic logical properties of relations and BoxE is proven to be fully expressive. However, BoxE is not able to capture composition. To further support composition, ExpressE \cite{DBLP:journals/corr/abs-2206-04192} embeds relations as hyper-parallelograms in the virtual triple space $\mathbb{R}^{2d}$ where the relational patterns are characterized through the spatial relationship of hyper-parallelograms. 
To embed hierarchical structure in KGs, HAKE \cite{DBLP:conf/aaai/ZhangCZW20} maps entities into the polar coordinate system to model semantic hierarchies. On the other hand, MuRP \cite{balazevic2019multi} embeds KGs into hyperbolic space and significantly reduces the dimensionality. 
To further embed relational patterns such as symmetry, inversion and composition, AttH \cite{chami2020low} represents relations as hyperbolic isometries including hyperbolic rotations and reflections parameterized by Givens matrices. HyboNet \cite{chen2021fully} formulates the relational transformation matrices by a Lorentz transformation that is decomposed into a Lorentz rotation and a Lorentz boost.
HBE \cite{pan2021hyperbolic} proposes a KGE model with an extended Poincare Ball where the hierarchy-specific parameters are optimized on the polar coordinate system. 
Recently, some works \cite{cao2022geometry,wang2021mixed,xiong2022ultrahyperbolic} argue that embedding KGs in a single homogeneous curvature space, no matter the zero-curvature Euclidean space, negatively curved hyperbolic space or positively curved hyperspheric space, cannot faithfully model the complex structures of KGs. MuRMP \cite{wang2021mixed} proposes to embed KGs into a mixed curvature space, namely a product of Euclidean, spherical and hyperbolic space. However, they do not explicitly model the possible interactions between these spaces. GIE \cite{cao2022geometry} combines Euclidean, hyperbolic and hyperspherical spaces and models their interactions by learning these complex spatial structures interactively between these different spaces. 
DGS \cite{iyer2022dual} further distinguishes the representations of the instance-view entities and the ontology-view concepts as they have inherently quite different structures (i.e., instance-view entities are organized more cyclically while ontology-view concepts are organized more hierarchically ). 
UltraE \cite{xiong2022ultrahyperbolic}, on the other hand, considers a pseudo-Riemannian space, namely, pseudo-hyperboloid equipped with a pseudo-Riemannian metric. This is inspired by the fact that  pseudo-hyperboloid generalizes hyperbolic and spherical spaces. Besides, UltraE models relations by hyperbolic and spherical rotation, allowing the inference of complex relation patterns. 

\begin{table}[]
    \centering
    \resizebox{\linewidth}{!}{
    \begin{tabular}{c|cccccc}
    Method & uncertainty & relational & set theory & structural \\ 
    \hline
    KG2E & $\checkmark$ & $\star$  & $\star$ & $\times$  \\
    DiriE & $\checkmark$ & $\star$ & $\times$ & $\times$  \\
    BEUrRE & $\checkmark$ & $\star$ & $\star$ & $\times$ \\
    BoxE & $\checkmark$ & $\star$ & $\checkmark$ & $\times$  \\
    ExpressE & $\checkmark$ & $\checkmark$ & $\checkmark$ & $\times$  \\
    MuRP & $\times$ & $\star$  & $\times$ & $\star$ \\
    AttH & $\times$ & $\star$  & $\times$ & $\star$ \\
    MuRMP & $\times$ & - & - & $\checkmark$  \\
    GIE & $\times$ & $\star$  & $\times$ & $\checkmark$ \\
    UltraE & $\times$ & $\star$  & $\times$ & $\checkmark$ \\
    DGS & $\times$ & $\star$  & $\times$ & $\checkmark$ \\
    \hline
    \end{tabular}
    }
    \caption{Characteristics of different geometric embeddings for KGs. 
    $\star$ means "partially", $-$ means not available. }
    \label{tab:characteristics_KGE}
\end{table}



\subsection{ Ontology/Hierarchy Completion.} 

Ontology completion aims to predict subsumption relations between classes in ontologies whose logical structure is expressed in the ontology axioms. Ontology embedding aims at embedding classes and relations such that the logical structure and the deductive closure can be preserved \cite{DBLP:conf/ijcai/KulmanovLYH19}. 
ELEm \cite{DBLP:conf/ijcai/KulmanovLYH19} proposes to embed each class and instance as a $n$-ball with a learnable radius and models each relation as a translation of the $n$-ball. This method is not able to represent intersectional closure because that $n$-ball is not closed under intersection, and the translational relation modeling suffers from a key issue, namely, not able to model relations between classes with different sizes. BoxEL \cite{xiong2022faithful} and ELBE \cite{DBLP:journals/corr/abs-2202-14018} both consider boxes as the class embedding due to their advantages of modeling intersectional closure. One major difference is that ELBE still uses translation as relation embeddings while BoxEL replaces it with affine transformation similar to BEUrRE \cite{DBLP:conf/naacl/ChenBCDLM21}. Furthermore, different from ELEm and ELBE which do not distinguish classes and instances, BoxEL embeds entities as points while classes as boxes, which is consistent with the classical conceptual space model where points denote objects, and regions denote concepts \cite{DBLP:books/daglib/0006106}. Box$^2$EL\cite{jackermeier2023box} further improves BoxEL and ELBE by embedding roles as boxes, in a similar way like BoxE \cite{abboud2020boxe}.
All these methods are limited to the embedding of ontology expressed by the lightweight Description Logics EL++, where full concept negation is not expressible. For the embeddings of ontologies expressed by $\mathcal{ALC}$ description logic where full negation is of interest, \cite{o2021cone} embeds each class as an angular convex cone and models the negation as the polarity of cone, which is inspired by the Farkas' Lemma \cite{dax1997classroom}

Hierarchy completion can be viewed as a special case of ontology completion where only one hierarchical relation (e.g., \emph{is\_a}) exists and the graph can be viewed as a hierarchy, partial order set/lattice or directed acyclic graph (DAG). 
Hence, the modeling of transitivity is an essential part of hierarchy completion.
Order embedding \cite{vendrov2015order} represents entities as axis-parallel cones. Probabilistic box embedding \cite{vilnis2018probabilistic,dasgupta2020improving} extends it to axis-parallel boxes while \cite{zhangmodeling} further extend it to deal with directed graphs with cycles. Hyperbolic semantic cone \cite{ganea2018hyperbolic} combines the advantages of hyperbolic space and cones.


\subsection{Hierarchical multi-label classification.}
HMC aims to associate each instance with multiple labels that are hierarchically organized in a directed acyclic graph. 
HCGE, the first geometric embedding model for HMC, models the uncertainty of node representations using Gaussian embeddings and shows improvement on multi-label node classification. 
Recent works \cite{DBLP:conf/nips/GiunchigliaL20} demonstrate that enforcing the output of an HMC model to respect the hierarchy constraint can boost the performance of HMC. 
To this end, MBM \cite{patel2021modeling} embeds instances and labels as boxes so that the logical structure can be respected, i.e., if an instance has a label then it also has all of the parent labels. 
Similar to MBM, Box2Type \cite{DBLP:conf/acl/OnoeBMD20} considers an instance of HMC, fine-grained entity typing, and embed entity types as boxes. Some methods \cite{DBLP:conf/aaai/ChenHXCJ20,DBLP:conf/acl/ChenHXJ20} propose to do HMC in hyperbolic space but they do not consider regions as type embeddings and hence cannot preserve logical structures. HMI \cite{xionghyperbolic}, on the other hand, embed each label as a curved hyperbolic hyperplane that corresponds to an enclosing $n$-ball and model the relationships between labels as geometric relations between the corresponding $n$-balls. HMI is able to preserve logical structure including both implication and exclusion. \cite{dhall2020hierarchical} leverage hyperbolic cones as label embeddings that aims to preserve label hierarchy. 

\subsection{Logical Query Answering.} 
Answering complex logical queries, for example, \emph{"list the countries that have hosted the Olympic Games but not the World Cup"}, is an important yet challenging task due to KGs incompleteness and the difficulty in properly modeling logic operations. Geometric and probabilistic query embedding methods provide a way to tractably handle logic operators in queries and equip excellent computing efficiency. Essentially, existing query embedding models represent logical queries by either geometry shapes or distributions. Logical operations, such as conjunction, disjunction, and negation, are performed as geometric or algebraic manipulations on them. Query2Box \cite{ren2019query2box} is the earliest work that maps each entity as a point and each query as a box. By doing so, this method models the conjunction of two query embeddings as the intersection of two boxes and evaluates the plausibility of each candidate answer based on the distance between the query box embedding and each entity point. Query2Box is limited in modeling negation due to the non-closure property of negation set in Euclidean space, i.e. the negation of a box embedding in Euclidean space is not a box. BetaE \cite{ren2020beta} remedies this problem by modeling queries as Beta distributions. It models the conjunction by computing the weighted product of beta distributions, and the negation by taking the reciprocal of the shape parameters $\alpha$ and $\beta$. Alternatively, ConE \cite{zhang2021cone} models conjunction and negation by representing queries as cones, i.e. the conjunction is the intersection of cones and the negation is the closure-complement. All of the above methods are not able to directly handle disjunctions, but can only transform the entire query into Disjunctive Normal Form (DNF) and perform the disjunction in the last step. GammaE \cite{yang2022gammae} and PERM \cite{choudhary2021probabilistic} enclose all logical operations in distributional (Gamma and Gaussian) spaces and allow them to be handled independently. Specifically, PERM encodes entities as Multivariate Gaussian distributions and queries as a mixture of Gaussian distributions. GammaE represents entities and queries using Gamma distributions. Similar to BetaE, these two methods also model the conjunction of query embeddings by the product of input distributions. However, both GammaE and PERM get rid of the DNF technique and model disjunction as the weighted mixture of Gamma and Gaussian distributions respectively.

\section{Summary and Future Work}
This paper conducts a comprehensive survey on geometric embedding methods for modeling relational data. Methods are systematically classified, discussed, and compared according to the proposed taxonomy. As we discussed, different geometric embeddings provide different inductive biases that are suitable for different tasks. We believe our comparison and analysis will shed light on more applications of geometric embeddings on more relational reasoning tasks. 

Despite its success, there are still challenges to be solved in this research field. We, therefore, suggest some promising directions for future research as follows
\paragraph{Heterogeneous hierarchies} Most current geometric embedding methods can only encode one hierarchy relation (e.g., \emph{is\_a}). However,  a real-world KG might simultaneously contain multiple hierarchical relations (e.g., \emph{is\_a} and \emph{has\_part}) \cite{patel2020representing}.
An embedding model should be able to distinguish these different hierarchical relations, especially when they are intertwined with each other. 
\paragraph{Deep geometric embeddings} Most current geometric relational embeddings are non-parametric models that do not require neural network parameterization. 
Developing deep architectures for current geometric relational embeddings is an exciting topic of future research.
\paragraph{Learning with symbolic knowledge} Incorporating symbolic knowledge, e.g., logical constraints over labels, into machine learning is in general a very promising direction as it enhances the robustness and interpretability of machine learning models. Geometric embeddings provide rich inductive biases for representing such logical constraints and have great potential for doing this. Besides modeling hierarchical constraints as in HMC, more kinds of logical constraints such as exclusion and intersection equivalence are worth exploring.







\bibliographystyle{named}
\bibliography{ijcai23}

\end{document}